\documentclass[10pt,twocolumn,letterpaper]{article}

\usepackage[pagenumbers]{cvpr} 

\usepackage{graphicx}
\usepackage{amsmath}
\usepackage{amssymb}
\usepackage{booktabs}

\usepackage{algorithm2e}
\usepackage[pagebackref,breaklinks,colorlinks]{hyperref}

\usepackage[capitalize]{cleveref}
\crefname{section}{Sec.}{Secs.}
\Crefname{section}{Section}{Sections}
\Crefname{table}{Table}{Tables}
\crefname{table}{Tab.}{Tabs.}

\def\cvprPaperID{10} 
\def\confName{CVPR}
\def\confYear{2023}

\begin{document}

\title{Online Lane Graph Extraction from Onboard Video  }

\author{ Yigit Baran Can\textsuperscript{1}\space\space\space\space Alexander Liniger\textsuperscript{1}\space\space\space\space Danda Pani Paudel\textsuperscript{1}\space\space\space\space Luc Van Gool\textsuperscript{1,2}\\
\textsuperscript{1}Computer Vision Lab, ETH Zurich\space\space\space\space \textsuperscript{2}VISICS, ESAT/PSI, KU Leuven \\ {\tt\small $\{$yigit.can, alex.liniger, paudel, vangool$\}$@vision.ee.ethz.ch} }
\maketitle

\begin{abstract}

Autonomous driving requires a structured understanding of the surrounding road network to navigate. One of the most common and useful representation of such an understanding is done in the form of BEV lane graphs. In this work, we use the video stream from an onboard camera for online extraction of the  surrounding's lane graph. Using video, instead of a single image, as input poses both benefits and challenges in terms of combining the information from different timesteps. We study the emerged challenges using  three different approaches. The first approach is a post-processing step that is capable of merging single frame lane graph estimates into a unified lane graph. The second approach uses the spatialtemporal embeddings in the transformer to enable the network to discover the best temporal aggregation strategy. Finally, the third, and the proposed method, is an early temporal aggregation through explicit BEV projection and alignment of framewise features. A single model of this proposed simple, yet effective, method can process any number of images, including one, to produce accurate lane graphs. The experiments on the Nuscenes and Argoverse datasets show the validity of all the approaches while highlighting the superiority of the proposed method. The code will be made public.  
\end{abstract}

\section{Introduction}

Autonomous driving requires accurate road scene understanding as a base for downstream tasks such as predicting the motion of agents~\cite{cui2019multimodal, hong2019rules, rella2021decoder, zaech2020action} and planning the ego-motion~\cite{DBLP:conf/rss/BansalKO19, chen2020learning}. One of the most important components of the road scene understanding is the extraction of the local road network. While there has been significant progress in offline HD-Maps~\cite{jaritz20202d,seif2016autonomous,ma2019exploiting,ravi2018real,casas2021mp3}, online road network extraction has been sparsely studied. The traffic scene is a highly dynamic environment and the road network frequently undergoes spontaneous changes that require immediate adaptation of the autonomous vehicle. Therefore, while offline HD-Maps are immensely useful, online lane graph estimation is a necessity. Furthermore, depending solely on offline maps severely limits the geographical areas autonomous driving. 

\begin{figure}
    \centering
    \includegraphics[width=\linewidth]{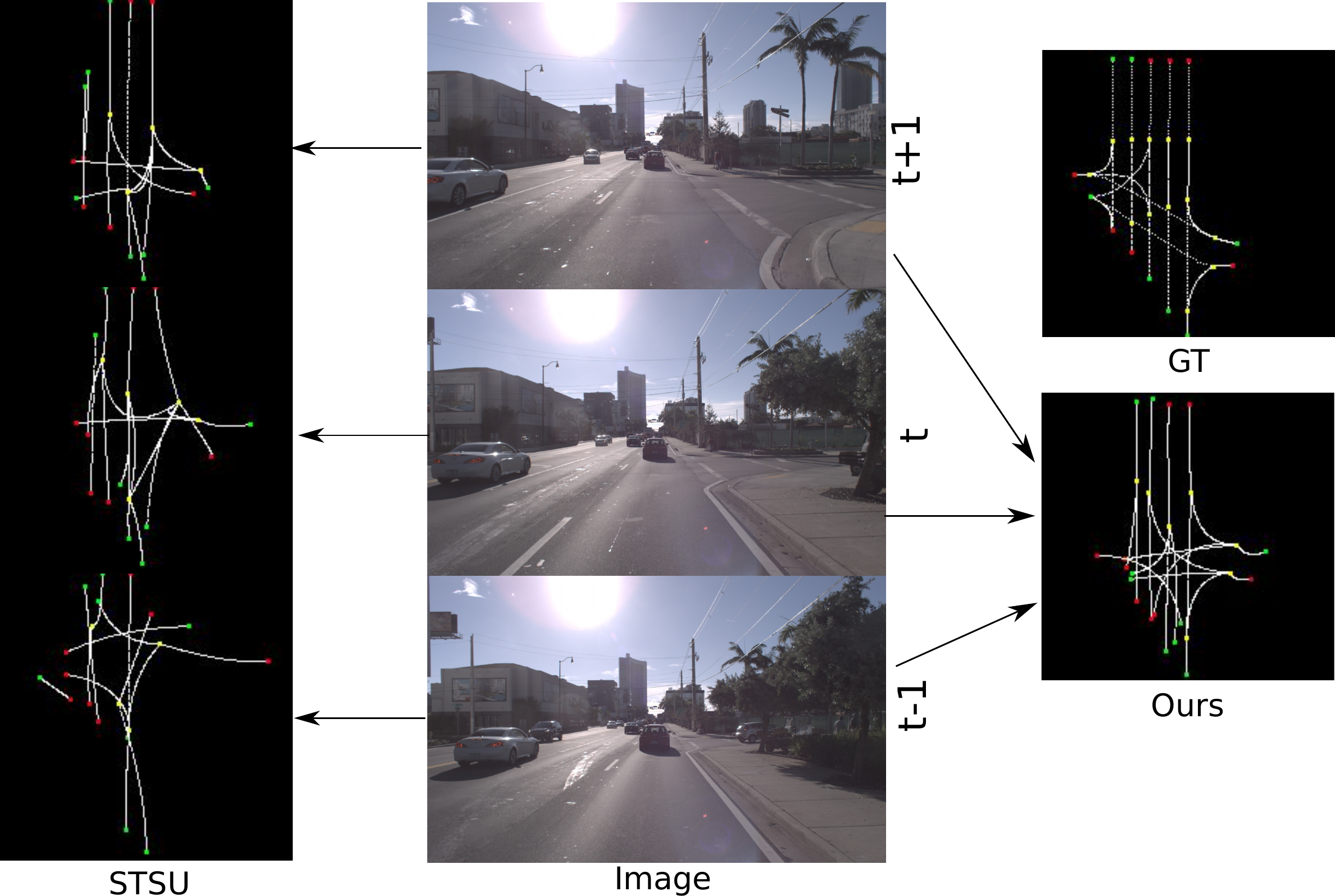}
    \caption{
   The existing methods such as STSU \cite{Can_2021_ICCV} only utilize a single frame while our method can combine information from several frames from a video to predict accurate lane graph. The ground truth (GT) lane graph is shown on the top-right. The reported result is obtained by the proposed suggested method.}
    \label{fig:teaser}
    
\end{figure}

One of the main challenges in online traffic scene understanding is the mismatch between the input and output planes. The input is one, or several, onboard camera image where the image plane is perpendicular to the output ground plane. Therefore, accurately understanding on the output Bird's-Eye-View (BEV) ground plane from the input image plane representation requires special care. Several works in literature have shown that image-plane based processing followed by the projection to BEV plane  results an inferior performance, when compared to to early projection techniques~\cite{DBLP:conf/cvpr/RoddickC20,can2020understanding,wu2020motionnet,paz2020probabilistic,yang2018hdnet,casas2018intentnet,philion2020lift, zhou2022cross}. Most of the existing works, however, produce semantic segmentation outputs on a BEV grid for general scene understanding instead of generating a lane graph required for the task of navigation. In this context, some methods output lane boundaries~\cite{DBLP:conf/ivs/NevenBGPG18, DBLP:conf/iccvw/GansbekeBNPG19}, while~\cite{homayounfar2019dagmapper} that extract lane graphs focus solely on highways and uses LiDAR data. This method largely depends on the existence of regular road markings and thus cannot be applied to urban or rural settings. Recently, an online traffic scene understanding method has been proposed~\cite{Can_2021_ICCV} that can extract local road networks from a single onboard camera image. The method works in a variety of settings and outputs a directed graph that represents the complete lane graph including the traffic flow directions. This type of output is suitable for the downstream use~\cite{liang2018end,paz2021tridentnet,liang2019convolutional,homayounfar2019dagmapper}. Later, the authors improve upon their results by considering minimal cycles that arise from the intersections of the centerlines~\cite{can2022topology}. The resulting method improves topological understanding and boosts the performance. However, this method requires computationally expensive pre-processing and the training is difficult. While the single frame setting has been explored, there has been no method that can produce a similar online lane graph outputs from an onboard video stream, up to our knowledge. Utilizing only a single image is inefficient since it discards the information from other frames that are overlapping and complimentary to the requested BEV understanding. Having several frames from different times allows a more informed prediction and alleviates occlusion problems. However, it also brings challenges to efficiently aggregate the information from different frames for accurate predictions.

In this work, we propose there approaches to combine temporal information in the described settings. Two of these approaches are developed to
benchmark the video-based lane graph estimate using strong baselines, while the third method is the proposed solution for the task at hand. In this context, at first, given SOTA single image based lane graph estimations, it stands to reason that most intuitive solution is to combine the individual frames' estimates in a  \emph{temporal post processing} step. To this end, we propose an algorithm that matches the centerlines of given lane graphs and updates the Bezier control points of the reference frame's centerlines iteratively. The second approach involves using a \emph{spatial-temporal embedding transformer} for lane graph estimation from video frames. This approach is motivated by~\cite{DBLP:conf/eccv/CarionMSUKZ20} where the temporal embedding indicates the timestep of feature maps from different frames. The spatial-temporal transformer network explores the best combination of the temporal information to produce the desired outputs. The final, and the suggested method proposed in this work, is a simple yet effective early temporal aggregation of feature maps in the BEV reference frame through explicit projection to the ground similar to~\cite{can2020understanding}. When the ground is assumed to be flat and horizontal, the suggested method is simple and straightforward which effectively learns to produce accurate estimates, despite of errors in the projections from image plane to the BEV. By aggregating the features in the BEV, the method is independent of the number of frames, and the same model instance can produce outputs with any number of input frames, including one. Similar explicit warping technıques have been used before in segmentation setting \cite{zhou2022cross, can2020understanding, DBLP:conf/cvpr/RoddickC20, philion2020lift} to project image level features on BEV. In this work, we also show how to use a similar aggregation method to train a transformer to operate on a temporally aggregated representation to output structured lane graph. Our experiments show that all proposed methods improve results over single frame estimates while the suggested (the proposed third method) temporal BEV aggregation method outperforms other methods and produces competitive results even when using a single frame.
One instance of such result is illustrated in Figure~\ref{fig:teaser}.

The major contributions of this paper lies on processing video input for the online lane graph extraction. In this process, we explore several aspects, whose main contributions can be summarized as follows:
\begin{enumerate}
\setlength{\itemsep}{0pt}
\setlength{\parskip}{0pt}
\item We propose three methods that utilize multiple video frames for accurate lane graph estimation. All three methods are superior to SOTA single frame based methods, two of which serve for benchmarking video-based lane graph extraction as strong baselines.
The third proposed method is the suggested method for use.

\item Ablations are carried out to show the flexibility of our method with respect to input frame configurations.
\item The results obtained by our method are superior to the compared baselines and the state-of-the-art methods even when using a single frame.

\end{enumerate}

\section{Related Work}

The literature on lane graphs extraction mostly focus on offline methods. The inputs to these methods are aerial images, LiDAR or some aggregated sensor data. Some earlier works on road network extraction focus on aerial images as the input data~\cite{auclair1999survey, richards1999remote}. Building on these works, over the years, significant improvements have been achieved in aerial image based lane graph extraction~\cite{batra2019improved,sun2019leveraging,ventura2018iterative}. HD-Maps provide a more detailed and rich representation of the traffic scene than only lane graphs. However, this rich structure is usually extracted offline. The input data for these methods can be both 2D (images) or 3D (LiDAR, radar,  etc.). A frequently applied technique in offline mapping is 3D data aggregation over several passes which enables the sparse 3D scans to be pooled in a much denser representation~\cite{liang2019convolutional, homayounfar2018hierarchical,liang2018end}. While offline HD-Maps are very useful, they still require accurate ego-vehicle localization since the offline maps are in a canonical frame.

Especially relevant to our work is the literature on structured representations of the traffic scenes. One set of these methods focus on extracting lane boundaries. Recently,~\cite{DBLP:conf/cvpr/HomayounfarMLU18} proposed a method to detect lane boundaries on highways where the boundaries are parameterized in the form of polylines. This parameterization allows for a recursive method to output the location of a control point one step at a time. Another work on a similar concept is~\cite{DBLP:conf/cvpr/HomayounfarMLU18} where the authors propose to use an RNN to estimate the initial boundary points using a 3D point cloud. These initial points form the initial state of a modified Polygon-RNN~\cite{DBLP:conf/cvpr/AcunaLKF18}. Polygon-RNN is then used to estimate a polyline. The common theme among these methods is their dependence on dense 3D clouds and the fact they are restricted to relatively simple highways. These methods cannot adapt to more complicated urban settings with camera inputs~\cite{Can_2021_ICCV}.

Another relevant part of literature is the task of lane estimation. In this comparison we focus on online lane estimation using an onboard monocular camera since it is the closest to our setting and the problem has been studied extensively in the literature~\cite{DBLP:conf/ivs/NevenBGPG18, DBLP:conf/iccvw/GansbekeBNPG19}. While some methods focus on extracting the lanes in the image plane~\cite{DBLP:conf/iccv/GarnettCPLL19, DBLP:journals/corr/abs-2002-06604}, others propose methods to project the information in image plane to BEV~\cite{DBLP:journals/corr/abs-2011-01535, DBLP:conf/siu/YenIaydinS18,DBLP:conf/ivs/NevenBGPG18}. However, these methods are shown to work only in simple cases such as highways and intersections and roundabouts pose a significant problem to them. Recently,~\cite{DBLP:conf/icra/LiWWZ22} proposed an online HD-Map method that builds upon~\cite{Can_2021_ICCV}. Their method does not produce a lane graph and outputs lane boundaries. This representation still does not provide traffic direction and connectivity between lanes, thus, requires a separate lane graph method such as~\cite{Can_2021_ICCV} to complete scene understanding. 

\begin{figure*}
    \centering
    \includegraphics[width=1\linewidth]{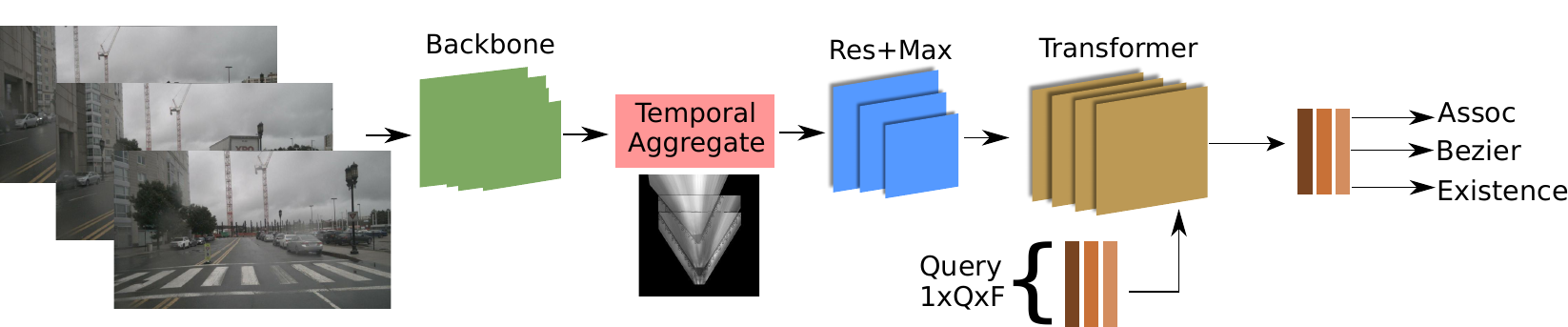}
    \caption{Overall architecture depends on the temporal aggregation module to combine information from multiple frames into a single BEV feature map. This feature map is then processed by a transformer to produce the lane graph estimate. The temporal aggregation step allows us to make predictions efficiently from the varying number of input frames. On the other hand, the transformer-based architecture serves best to estimate the structured outputs in the form of the lane graphs. 
    }
    \label{fig:arch}
\end{figure*}

BEV representations of the traffic scene mostly focus on semantic segmentation. In this context, some methods use camera images~\cite{DBLP:conf/cvpr/RoddickC20,philion2020lift,can2020understanding}, some other methods combine camera with LiDAR information~\cite{pan2020cross,hendy2020fishing}. However, the output representation for these methods, while useful, still require further processing to be used by the downstream tasks. The first online structured traffic scene estimation method was proposed in~\cite{Can_2021_ICCV} where the authors propose several baselines apart from their proposed method. The lane graph representation is a directed graph with the centerlines being represented by the directed graph vertices. The edges of the graph show whether two given centerlines are connected. Later, authors improve their results through explicitly modelling topology~\cite{can2022topology}. However, these methods only use a single frame which can be perceived as inefficient since the information from the video stream is not fully utilized, which we tackle in this paper.

\section{Method}

\subsection{Lane Graph Representation}
\label{lanegraph}

Lane graphs are directed graphs that represent the road network taking the traffic direction into consideration. The vertices of the graph represent the centerlines and the edges encode the ``connectivity'' of the centerlines. The graph $G(V, E)$ has $V$ as the vertices and the edges $E \subseteq \{(x,y)\; |\; (x,y) \in V^2\}$. The incidence matrix $A$ can represent the edges. A centerline $x$ is connected to another centerline $y$, i.e. $(x,y) \in E$ if and only if the centerline $y$'s starting point is the same as the end point of the centerline $x$. This means $A[x,y] = 1$ if the centerlines $x$ and $y$ are connected. Following~\cite{Can_2021_ICCV}, we represent the centerlines, vertices of the graph $G$, with Bezier curves. This lane graph representation is shown to be an effective way to outline the scene information on complicated scenes such as crossroads or roundabouts. 

\subsection{Overview}

Out of the three methods we devise in this paper, the early temporal aggregation method is presented as the proposed method while the others are examined as baselines. Our goal is creating a method that can accept any number of input frames that are arbitrarily distributed temporally. This ensures the flexibility of the method to work with a single frame, in a causal setting (only past frames) or using past and future frames for delayed outputs. Let us assume that there are $N$ RGB input images $Im$ of dimensions $N\times H\times W \times 3$ which are processed by some deep neural network backbone to produce image plane feature maps, $X_n$ of dimensions $N\times H_X\times W_X \times F$, for each input image separately. These features then get warped to the BEV using a homograph and a flat road assumption. Given the pose between the images, the BEV feature maps can be aggregated. Therefore, we define the \textbf{target} bBEV area in which we will estimate the lane graph, be of dimensions $H' \times W'$. This area is centered in lateral direction (W') on ego-vehicle and extends in longitudinal dimension from the ego-vehicle (H'). Let us also define an extended area that completely includes target area. This extended area is called field-of-view (\textbf{FOV}) which is of dimensions $H'' \times W''$. Both the target and the extended FOV share a common resolution $Rs$ where $H'(W')\times Rs$ converts the grid into metric units.

The temporally aggregate BEV feature maps areprocessed by a transformer~\cite{DBLP:conf/eccv/CarionMSUKZ20} based module similar to~\cite{Can_2021_ICCV, can2022topology}. Each learnt query vector outputs the centerline Bezier control points, existence probability and an association feature vector which is used to determine the connectivity. Compared to \cite{Can_2021_ICCV, can2022topology} the main difference in the transformer module is the input. In our paper we warp and temporally aggregate the features in the BEV compared to directly processing the image feature maps. Using the BEV feature not only allows for our efficent temporal aggregation but also alleviates the transformer from reason about the view change. In the next sections, we will explain the warping and temporal aggregation followed by the proposed baselines.


\subsection{Architecture}
Given $N$ frames, the process starts with obtaining feature maps from a backbone network. The image level feature maps are warped to ground plane and aligned using the known camera poses. The aligned feature maps are temporally aggregated, see next section for more details. The aggregated BEV feature map is of dimensions $1\times H'' \times W'' \times F$ where $F$ is the number of feature channels. The resulting representation is processed further by a convolutional neural network to reduce the spatial size through a series of max pooling and residual layers, see Fig~\ref{fig:arch}. Finally a transformer based network produces centerline estimates. These estimates are represented by learnt query vectors that attend to the input to produce 3 outputs: a) the control points for the Bezier curve that represents a centerline, b) the probability of existence for that centerline, c) the association feature vector that are used to predict the connectivity of the lane graph $G$. Note that the split BEV-Image level positional embedding employed in \cite{Can_2021_ICCV, can2022topology} is rendered unnecessary in our formulation since the feature map is explicitly projected to the target BEV plane. Our experiments show the impact this change makes through the performance of the proposed method when using single frame.

\subsection{Temporal Aggregation}

Let us represent the reference frame with $Im_R$. The reference frame defines the time step of interest, thus the target BEV area. Our goal is to transform the information from the backbone processed frames, $X_n$, into the BEV target region. To this end, for each frame $X_n, n \in N$, we have a pose matrix $P_n$ that represents the position and pose of the ego-vehicle in global coordinate system, a camera extrinsics matrix $C_n$ that represents the position and pose of the camera with respect to the ego vehicle and, finally, a camera intrinsics matrix $M_n$. In real world scenario, these matrices can be obtained using the odometry pipeline on the vehicle. 
The resulting warped feature map is denoted by $Y_n$.

The projective homography applied on the feature map can be described as a spatial resampling of the feature map. For the reference frame,
\begin{align}
    Y_R[C_R^{-1} M_R^{-1}(i,j,1)] = X_R[i,j], \quad i,j \in [H_X,W_X]\,,
    \label{eq:projectref}
\end{align}
where $(i,j,1)$ denotes the homogeneous coordinates of the image plane feature map indices. In order to obtain the feature vectors corresponding to a grid location $(m,n) \in [H'',W'']$, we use bilinear interpolation.

In order to warp the feature maps of the non-reference frames, we move the ground projected feature map to the reference frame's global coordinate position, i.e.,
\begin{align}
    Y_n[P_R P_n^{-1}C_n^{-1} M_n^{-1}(i,j,1)] = X_n[i,j], \quad i,j \in [H_X,W_X]\,.
    \label{eq:projectother}
\end{align}
The warping operations are carried out with the assumption of a flat ground plane. Therefore, each image level feature is treated to be the 2D projection of a plane that has fixed vertical distance to the camera. This assumption is obviously violated by most of the pixels in practice. However, the later stages of the network learn to work with these violations to produce accurate outputs. 

\begin{figure*}
    \centering
    \includegraphics[width=\linewidth]{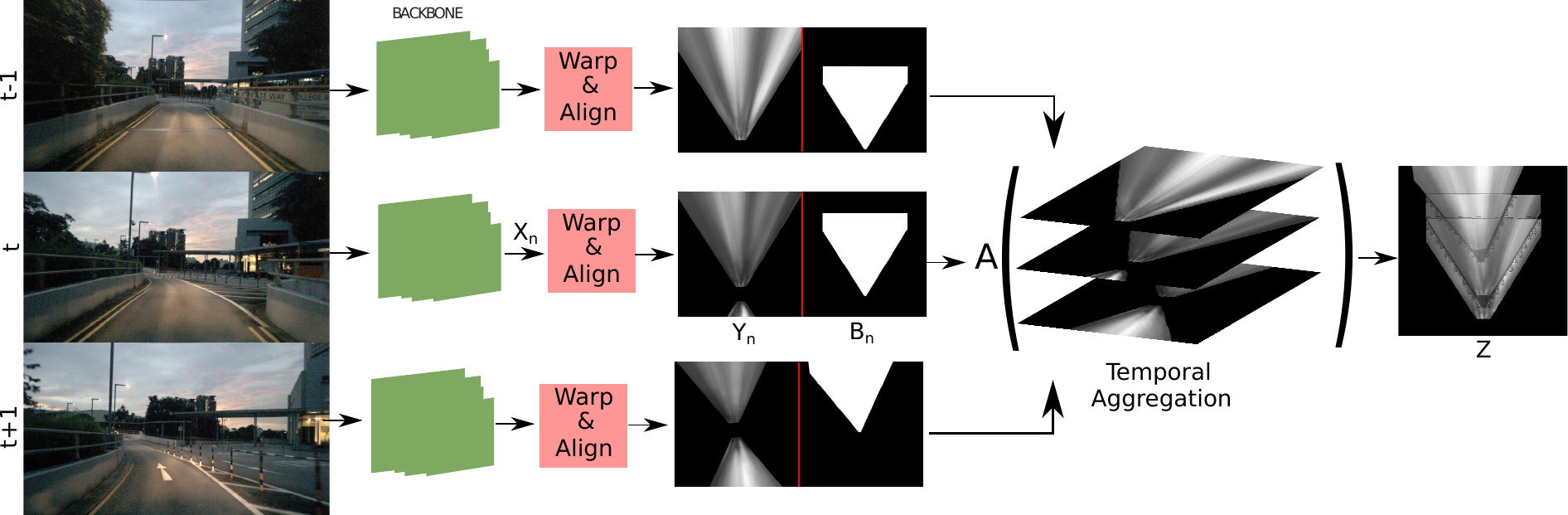}
    \caption{Temporal aggregation process converts backbone features from frames into one BEV feature map $Z$. To this end, each image level feature map is warped to ground level with flat ground plane assumption and then aligned into the target area in the reference frame's coordinate system. The warped and aligned feature maps are processed by the temporal aggregation function $A$, in this figure we show max operator, to produce final feature map $Z$.
    }
    \label{fig:temporalagg}
\end{figure*}

Projecting the image plane features to the ground results in a half open plane where the image pixels close to the focal point level are projected to infinity while the upper part of the image is wrapped around and manifest itself behind the ego-vehicle's position in the BEV. In order to eliminate these artifacts, we mask the warped features $Y_n$ with a binary BEV mask $B_n$. BEV mask covers the same area as the target area and it is warped similar to Eq. \ref{eq:projectref} and \ref{eq:projectother}.  

The warped and aligned feature maps from temporally distributed frames that need to be aggregated to be further processed. Since our goal is to allow for any number of input frames, we experiment with framewise \textbf{mean} or \textbf{max} aggregation functions. In our experiments, we observed that \textbf{max} operation performs better than \textbf{mean}. Referring to an aggregation function as $A(.)$, we can write the aggregated feature $Z$ with dimensions $H'' \times W''$ as 
\begin{align}
    Z_{ij} = A(B_{t,ij}Y_{t,ij}), \quad i,j \in [H'',W'']\,.
    \label{eq:finfeat}
\end{align}
The whole process is visualized in Figure~\ref{fig:temporalagg} where one future and one past frame are used to produce the feature map $Z$ that corresponds to the reference frame $t$. In practice, we first process the warped frames with a residual block and then apply max operation. Since the warped frames are stacked in the batch dimension, processing the warped frames before max operation does not put any restrictions on the number of frames.

\section{Baselines}
In this section, we explain the other methods that we devised to merge multi frame information. The first method aims to aggregate the information directly in the transformer  through spatial-temporal embeddings (STETR). The second approach is a post-processing method that only works on the final per-frame estimations.

\subsection{Spatial-Temporal Embedding Transformer}

Given the image level feature maps $X_n$ of the frames, our proposed method aggregates these feature maps in the spatial dimension to create one feature map. However, it is possible to use all the feature maps as they are by indicating their temporal location to the network. To this end, we use the same transformer architecture as~\cite{Can_2021_ICCV} with a small change in the positional embedding of the transformer. The positional embeddings provide the only information regarding the ordering in the transformers and when transformers are applied on images, the common positional embeddings encode 2D locations~\cite{DBLP:conf/eccv/CarionMSUKZ20, can2022topology}. When transformers are applied on video input, the resulting spatial-temporal nature of the features are embedded in the positional encodings~\cite{DBLP:conf/eccv/YuMRZY20, DBLP:journals/corr/abs-2109-12218, DBLP:conf/cvpr/DukeA0AT21}. In a similar way, we extend the spatial 2D positional embedding of~\cite{Can_2021_ICCV} with temporal information. The temporal embedding is represented with $E(t)^{(i)}$ where $i$ indexes along the feature dimension and $t$ represents the relative temporal position of the frame with respect to the reference frame. The relative position is simply the difference between the frame number of the current frame and the reference frame. Note that, this means that the past frames have a negative relative temporal position. Then the embedding $E(t)^{(i)}$ is given by:
\begin{align}
      E(t)^{(i)}= 
\begin{cases}
    \text{sin}(\omega t),& \text{if } i=2k\\
   \text{cos}(\omega t),& \text{if } i=2k+1\\
\end{cases}\,,
    \label{eq:tempembed}
\end{align}
where $k$ is an integer and $\omega$ is a fixed frequency hyper-parameter. We use the same frequency hyper-parameter as the original DETR \cite{DBLP:conf/eccv/CarionMSUKZ20} implementation. 

The image level feature maps $X \in \mathbb{R}^{N\times X\times Y\times F}$ are flattened to a single sequence of dimensions $1\times NXY\times F$ which is processed by the transformer. We call the resulting baseline method STETR.  

\subsection{Temporal Post Processing}

As a second baseline, we propose to merge the lane graph estimates that are obtained from single frame inputs. To this end, let us represent the lane graph estimate from the frame $n$ with $G_n$. Then, the Bezier coefficient estimates of the frame $n$ are denoted by $R_n \in \mathbb{R}^{Q\times 3\times 2}$, the estimated existence probability for the centerline candidates by $P_n \in \mathbb{R}^{Q\times 1}$ and the connectivity estimates by $C_n \in \mathbb{R}^{Q\times Q}$. Moreover, let us represent the interpolated centerlines with $\Omega_n$ with dimensions $Q\times \omega \times 2$ where $\omega$ is the number of interpolation points which is 100 in our paper. 

The algorithm is outlined in Algorithm~\ref{alg:cap}. The central idea is to match the centerline estimates from the non-reference frame to the centerline estimates in the reference frame. However, this is a very challenging task since the network estimates for the same centerline among different frames are not consistent and frequently fragmented. Therefore, we use three criteria in matching the centerlines. First of all, we only operate on the centerline estimates that have existence probability above a threshold prob\_thresh. Then, in order to differentiate between centerlines that are close by in opposite traffic direction, we obtain a direction vector that is is simply a unit vector from the initial Bezier control point to the last control point. For each centerline estimate in reference frame, the centerline estimates in other frames are checked if their direction vector has a dot product over some threshold dir\_thresh. For the centerline estimates that satisfy direction criterion, the candidates, the final test is the average distance of the interpolated points. Here, we require that there exists some candidate point that is closer than a threshold dist\_thresh to at least half of the reference centerline points. If all these criteria are met, the control points of the reference centerline are updated depending on the comparative distance of the initial and the final candidate control point to the reference centerline.

 


\begin{algorithm}
\caption{Temporal Post-processing algorithm. }\label{alg:cap}

$I_n\gets P_n \geq$ prob\_thresh\;

 \For{$i\gets0$ \KwTo $N$}{
    $R_n\gets R_n[I_n]$\;

    $P_n\gets P_n[I_n]$\;
    
    $C_n\gets C_n[I_n,I_n]$\;
    
    $D_n\gets \text{normalize}(R_n[:,-1]-R_n[:,0])$\;
    
    }
    
    $D_o\gets [D_n] \text{for}\quad n \neq r$\;
    
    $R_o\gets [R_n] \text{for}\quad n \neq r$\;
    
    $\Omega_o\gets [\Omega_n] \text{for}\quad n \neq r$\;

 \For{$i\gets0$ \KwTo $Q$}{
 
 $\text{dir\_sel} \gets D_r[i] \cdot D_o^T > \text{direction\_thresh}$ \;
 \For{$j\gets0$ \KwTo $(N-1)Q$}{
 \If{dir\_sel[j]}
    {
        $\text{inter\_dist}\gets|\Omega_r[i] - \Omega_o[j]| < \text{dist\_thresh}$ \;
        $\text{any\_dist}\gets\text{any}(\text{inter\_dist},\text{dim=1})$ \;
 $\text{dist\_sel} \gets \text{sum}(\text{any\_dist}) > 0.5\omega$ \;
 
  \If{dist\_sel}
    {
    $var1 \gets \text{min}(\text{inter\_dist}[0])$ \;
 $var2 \gets \text{min}(\text{inter\_dist}[-1])$ \;

    \eIf{$var1 \geq var2$}
    {
    $R_r[i]\gets [R_r[i][0], R_o[j][1:]]$\;
     }{
     $R_r[i]\gets [R_o[j][:-1],R_r[i][-1]]$\;
     }
 }
    }
 
    }

    }

\end{algorithm}



\section{Experiments}

We use the NuScenes~\cite{nuscenes2019} and Argoverse~\cite{DBLP:conf/cvpr/ChangLSSBHW0LRH19} datasets. The data processing and train/test splits are the same as~\cite{Can_2021_ICCV}. 

\noindent\textbf{Implementation.}
The BEV target area that our method outputs the lane graph for is from -25 to 25m in x-direction and 1 to 50m in z direction with a 25cm resolution. 
Our implementation is in Pytorch and runs with 10FPS with a single frame and 3FPS with 3 frames. The network is trained with 3 frames (1 past and 1 future). The time difference between frames is set to 2 seconds for both datasets for testing. During training, we sample frames in a 4 seconds range, which acts as a data augmentation. For the temporal aggregation function, we use the 'max'. The image backbone network is Deeplab v3+ \cite{DBLP:conf/eccv/ChenZPSA18} pretrained on Cityscapes dataset \cite{Cordts2016Cityscapes}.

\noindent\textbf{Baselines.}
We compare against state-of-the-art~\textbf{STSU}~\cite{Can_2021_ICCV}, \textbf{TPLR}~\cite{can2022topology}, and our proposed baselines where \textbf{STETR} refers to the Spatiotemporal Embedding Transformer and \textbf{PostMerge} refers to the proposed post processing method that is applied to~\textbf{STSU estimates}. The numbers in parentheses show the number of frames used.

\noindent\textbf{Ablations.} We carried out our ablations on the Nuscenes dataset. The performance of a \textbf{single model} of our proposed model is tested with different configurations of input frames.

\section{Results}

We report the comparison against SOTA in two benchmark datasets. In Tab~\ref{tab:nusc}, we present the results in NuScenes dataset. The first observation is that the PostMerge method provides consistent improvement over STSU showing that our post-processing works and that it is a good baseline. Moreover, we see that STETR produces competitive results to existing SOTA but fails to provide the best results. STETR performs much better than post processing or single frame SOTA methods in connectivity metric. This is expected since the connectivity metric requires the joint processing of the centerlines and post-processing does not provide this. Another important observation is that the proposed method outperforms all SOTA as well as baselines using a single frame. This shows that explicit warping of the image level backbone features to Bird's-Eye-View provides better representation than providing the BEV indices of the image level features as done by STSU and TPLR. When more frames (1 past and 1 future) are used, our method produces even better results. This shows the proposed temporal aggregation method's effectiveness. 

The quantitative results on Argoverse dataset are given in Tab \ref{tab:argo}. The first observation is that PostMerge produces results comparable to TPLR. Our method with single frame outperforms the baselines in M-F score. Moreover, using multiple frames further pushes the performance in all metrics to provide the best results except detection metric. Similar to the results in NuScenes dataset, 
 it can be seen that connectivity metric requires joint processing. Overall, in both datasets, the proposed method produces the best results and provides a flexible way to incorporate multiple frames with arbitrary temporal locations.











\begin{table}[h]
\begin{center}

\tabcolsep=0.25cm
\begin{tabular}{ |c|c|c|c| }
\hline
& \multicolumn{3}{|c|}{NuScenes} \\
\hline
\textbf{Method} & M-F &  Detect & C-F  \\
\hline
\hline


STSU(1) \cite{Can_2021_ICCV}& 56.7  & 59.9 & 55.2 \\

TPLR(1) \cite{can2022topology} &  58.2 & 60.2  & 55.3 \\

PostMerge(3) & 57.3 & 59.9 & 56.0 \\

STETR(3) & 57.2 & 60.0 & 61.2\\

Ours(1) & 58.3 & 60.0 & 61.0 \\

Ours(3) & \textbf{58.8} & \textbf{60.3} & \textbf{61.3} \\

\hline
\end{tabular}

\end{center}
\caption{Results on NuScenes benchmark datasets. Our method performs better than all the compared methods. Number in parentheses show the number of frames used. For 3 frames, one past and 1 future frame, as well as the reference frame are used. PostMerge and STETR are the proposed baseline methods for benchmarking the video-based lane graph extraction using strong baselines.}
\label{tab:nusc}
\end{table}

\begin{table}[h]
\begin{center}

\tabcolsep=0.25cm
\begin{tabular}{ |c|c|c|c| }
\hline
&  \multicolumn{3}{|c|}{Argoverse} \\
\hline
\textbf{Method}  &  M-F &  Detect & C-F \\
\hline
\hline


STSU(1) \cite{Can_2021_ICCV}& 55.6 & 60.1 & 54.9\\

TPLR(1) \cite{can2022topology} &  57.1 & 64.2 & 58.1\\

PostMerge(3) & 57.1 & \textbf{64.2} & 58.0\\

STETR(3) &57.2 & 62.0 & 56.0\\

Ours(1) & 57.9 & 62.1 & 57.8\\

Ours(3) &  \textbf{58.2}& 62.2 & \textbf{58.1}\\

\hline
\end{tabular}

\end{center}
\caption{Results on Argoverse benchmark dataset. The proposed method significantly outperforms the SOTA. Number in parentheses show the number of frames used. For 3 frames, one past and 1 future frame, as well as the reference frame are used. PostMerge and STETR are the proposed baseline methods for benchmarking the video-based lane graph extraction using strong baselines.}
\label{tab:argo}
\end{table}

\begin{figure*}
    \centering
    \includegraphics[width=1\linewidth]{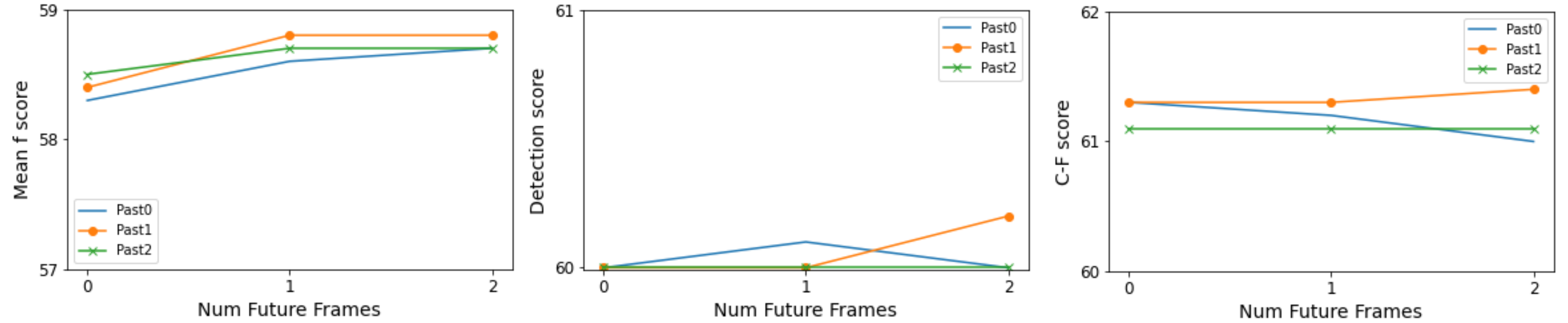}
    \caption{Performance of \textbf{a single model} of our proposed (suggested) method with respect to different number of past and future input frames. Each line shows the performance change given a fixed number of past frames with changing number of future frames. The importance of future frames is higher when compared to that of the past frames higher than one. 
    }
    \label{fig:graphs}
\end{figure*}

\begin{figure*}
    \centering
    \includegraphics[width=\linewidth]{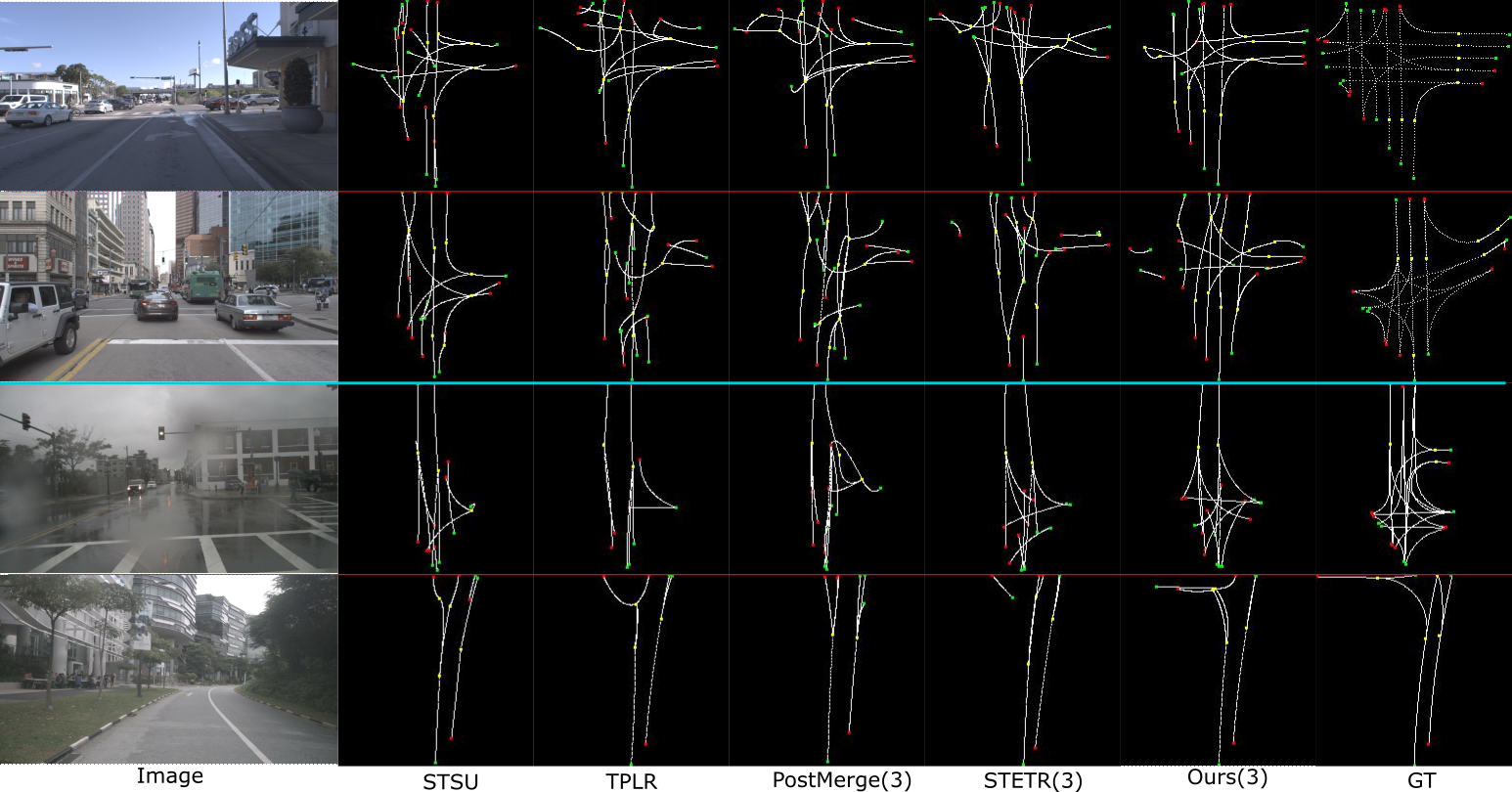}
    \caption{The visual results from Argoverse (top two rows) and Nuscenes (bottom two rows) benchmark datasets. The visuals confirm that the proposed (suggested) method produces more accurate lane graphs then the compared method and the established baselines. PostMerge and STETR are the proposed baselines for benchmarking the video-based lane graph extraction using strong baselines.
    }
    \label{fig:results}
   
\end{figure*}

In Figure~\ref{fig:graphs}, we provide the performance of our proposed method with respect to different number of past and future input frames. Note that it is always the same exact model checkpoint, with the only difference being the number of frames. The Mean-F score is steadily increasing as the number of past or future frames increases. We see that the detection score is somewhat independent of the input frame configuration. A similar trend also exists for the connectivity metric (C-F) where the addition of future frames seems to be helpful as long as there is at least one past frame. This can be the result of the training scheme where we always provide at least one past frame in our training procedure.

The visual results are given in Figure~\ref{fig:results}, which confirm the numerical results and indicate the superiority of the proposed method. The top two rows shows examples from Argoverse dataset while the bottom two rows are from NuScenes dataset. The single frame methods are consistently outperformed by the aggregation methods in both datasets. Apart from our proposed method, STETR produces the most visually accurate results. This indicates that the proposed baselines are also effective. Comparing our method with STETR, it can be seen that our method improves the performance especially in the occluded regions. For example, in the bottom two rows, the left turns in both examples are missed in STETR and other SOTA, while our method manages to capture these parts of the lane graph. A similar observation holds in the Argoverse samples where in the top example, our method four distinct lanes while the others fail to produce distinct centerlines.

\section{Conclusion}

In this work we address the task of extracting a complete lane graph using aggregating information from multiple frames from a single onboard camera. We propose three approaches to overcome the challenge of merging information from temporally distributed frames to produce a structured output. The post-processing based method (PostMerge) matches the lane graphs estimates from single frame methods. The matched lane graphs are used to iteratively update centerline control points. The second method (STETR) aims to provide the transformer based network information regarding the spatial-temporal locations of input feature map grid locations. The network, then, learns to combine multiple frames into a unified lane graph. The final, and suggested, method employs a simple yet effective early temporal aggregation through explicit projection of image level features into ground plane. This process aligns the representation plane and the output plane, allowing for the transformer to work with standard spatial position embeddings. Our experiments show that all three methods outperform single frame SOTA methods. We treat the proposed PostMerge and STETR as baselines for benchmarking the video-based lane graph extraction using strong baselines. The suggested method, which performs early fusion of the temporal information on the BEV plane performs better than the compared SOTA method and the established baselines. We believe, Our exhaustive study of the online extraction of land graph from video frames, with strong baselines and benchmarking, will open of new avenue for the future research directions.

\section{Limitations}
The proposed method requires relative localization of all input frames with respect to the reference frame. However, we produce estimates in the ego centered reference frame. Thus, an accurate localization in a global reference frame is not necessary and we only require the measurement of relative location changes in a short time window ($\sim$2s).

{\small
\bibliographystyle{ieee_fullname}
\bibliography{egbib}

\begin{thebibliography}{10}\itemsep=-1pt

\bibitem{DBLP:conf/cvpr/AcunaLKF18}
David Acuna, Huan Ling, Amlan Kar, and Sanja Fidler.
\newblock Efficient interactive annotation of segmentation datasets with
  polygon-rnn++.
\newblock In {\em 2018 {IEEE} Conference on Computer Vision and Pattern
  Recognition, {CVPR} 2018, Salt Lake City, UT, USA, June 18-22, 2018}, pages
  859--868. {IEEE} Computer Society, 2018.

\bibitem{auclair1999survey}
M-F Auclair-Fortier, Djemel Ziou, Costas Armenakis, and Shengrui Wang.
\newblock Survey of work on road extraction in aerial and satellite images.
\newblock 1999.

\bibitem{DBLP:conf/rss/BansalKO19}
Mayank Bansal, Alex Krizhevsky, and Abhijit~S. Ogale.
\newblock Chauffeurnet: Learning to drive by imitating the best and
  synthesizing the worst.
\newblock In Antonio Bicchi, Hadas Kress{-}Gazit, and Seth Hutchinson, editors,
  {\em Robotics: Science and Systems XV, University of Freiburg, Freiburg im
  Breisgau, Germany, June 22-26, 2019}, 2019.

\bibitem{batra2019improved}
Anil Batra, Suriya Singh, Guan Pang, Saikat Basu, CV Jawahar, and Manohar
  Paluri.
\newblock Improved road connectivity by joint learning of orientation and
  segmentation.
\newblock In {\em Proceedings of the IEEE/CVF Conference on Computer Vision and
  Pattern Recognition}, pages 10385--10393, 2019.

\bibitem{nuscenes2019}
Holger Caesar, Varun Bankiti, Alex~H. Lang, Sourabh Vora, Venice~Erin Liong,
  Qiang Xu, Anush Krishnan, Yu Pan, Giancarlo Baldan, and Oscar Beijbom.
\newblock nuscenes: A multimodal dataset for autonomous driving.
\newblock {\em arXiv preprint arXiv:1903.11027}, 2019.

\bibitem{Can_2021_ICCV}
Yigit~Baran Can, Alexander Liniger, Danda~Pani Paudel, and Luc Van~Gool.
\newblock Structured bird's-eye-view traffic scene understanding from onboard
  images.
\newblock In {\em Proceedings of the IEEE/CVF International Conference on
  Computer Vision (ICCV)}, pages 15661--15670, October 2021.

\bibitem{can2022topology}
Yigit~Baran Can, Alexander Liniger, Danda~Pani Paudel, and Luc Van~Gool.
\newblock Topology preserving local road network estimation from single onboard
  camera image.
\newblock In {\em Proceedings of the IEEE/CVF Conference on Computer Vision and
  Pattern Recognition}, pages 17263--17272, 2022.

\bibitem{can2020understanding}
Yigit~Baran Can, Alexander Liniger, Ozan Unal, Danda Paudel, and Luc Van~Gool.
\newblock Understanding bird's-eye view semantic hd-maps using an onboard
  monocular camera.
\newblock {\em arXiv preprint arXiv:2012.03040}, 2020.

\bibitem{DBLP:conf/eccv/CarionMSUKZ20}
Nicolas Carion, Francisco Massa, Gabriel Synnaeve, Nicolas Usunier, Alexander
  Kirillov, and Sergey Zagoruyko.
\newblock End-to-end object detection with transformers.
\newblock In Andrea Vedaldi, Horst Bischof, Thomas Brox, and Jan{-}Michael
  Frahm, editors, {\em Computer Vision - {ECCV} 2020 - 16th European
  Conference, Glasgow, UK, August 23-28, 2020, Proceedings, Part {I}}, volume
  12346 of {\em Lecture Notes in Computer Science}, pages 213--229. Springer,
  2020.

\bibitem{casas2018intentnet}
Sergio Casas, Wenjie Luo, and Raquel Urtasun.
\newblock Intentnet: Learning to predict intention from raw sensor data.
\newblock In {\em Conference on Robot Learning}, pages 947--956, 2018.

\bibitem{casas2021mp3}
Sergio Casas, Abbas Sadat, and Raquel Urtasun.
\newblock Mp3: A unified model to map, perceive, predict and plan.
\newblock {\em arXiv preprint arXiv:2101.06806}, 2021.

\bibitem{DBLP:conf/cvpr/ChangLSSBHW0LRH19}
Ming{-}Fang Chang, John Lambert, Patsorn Sangkloy, Jagjeet Singh, Slawomir Bak,
  Andrew Hartnett, De Wang, Peter Carr, Simon Lucey, Deva Ramanan, and James
  Hays.
\newblock Argoverse: 3d tracking and forecasting with rich maps.
\newblock In {\em {IEEE} Conference on Computer Vision and Pattern Recognition,
  {CVPR} 2019, Long Beach, CA, USA, June 16-20, 2019}, pages 8748--8757.
  Computer Vision Foundation / {IEEE}, 2019.

\bibitem{chen2020learning}
Dian Chen, Brady Zhou, Vladlen Koltun, and Philipp Kr{\"a}henb{\"u}hl.
\newblock Learning by cheating.
\newblock In {\em Conference on Robot Learning (CoRL)}, 2020.

\bibitem{DBLP:conf/eccv/ChenZPSA18}
Liang{-}Chieh Chen, Yukun Zhu, George Papandreou, Florian Schroff, and Hartwig
  Adam.
\newblock Encoder-decoder with atrous separable convolution for semantic image
  segmentation.
\newblock In Vittorio Ferrari, Martial Hebert, Cristian Sminchisescu, and Yair
  Weiss, editors, {\em Computer Vision - {ECCV} 2018 - 15th European
  Conference, Munich, Germany, September 8-14, 2018, Proceedings, Part {VII}},
  volume 11211 of {\em Lecture Notes in Computer Science}, pages 833--851.
  Springer, 2018.

\bibitem{Cordts2016Cityscapes}
Marius Cordts, Mohamed Omran, Sebastian Ramos, Timo Rehfeld, Markus Enzweiler,
  Rodrigo Benenson, Uwe Franke, Stefan Roth, and Bernt Schiele.
\newblock The cityscapes dataset for semantic urban scene understanding.
\newblock In {\em Proc. of the IEEE Conference on Computer Vision and Pattern
  Recognition (CVPR)}, 2016.

\bibitem{cui2019multimodal}
Henggang Cui, Vladan Radosavljevic, Fang-Chieh Chou, Tsung-Han Lin, Thi Nguyen,
  Tzu-Kuo Huang, Jeff Schneider, and Nemanja Djuric.
\newblock Multimodal trajectory predictions for autonomous driving using deep
  convolutional networks.
\newblock In {\em International Conference on Robotics and Automation (ICRA)},
  pages 2090--2096, 2019.

\bibitem{DBLP:conf/cvpr/DukeA0AT21}
Brendan Duke, Abdalla Ahmed, Christian Wolf, Parham Aarabi, and Graham~W.
  Taylor.
\newblock {SSTVOS:} sparse spatiotemporal transformers for video object
  segmentation.
\newblock In {\em {IEEE} Conference on Computer Vision and Pattern Recognition,
  {CVPR} 2021, virtual, June 19-25, 2021}, pages 5912--5921. Computer Vision
  Foundation / {IEEE}, 2021.

\bibitem{DBLP:journals/corr/abs-2011-01535}
Netalee Efrat, Max Bluvstein, Shaul Oron, Dan Levi, Noa Garnett, and Bat~El
  Shlomo.
\newblock 3d-lanenet+: Anchor free lane detection using a semi-local
  representation.
\newblock {\em CoRR}, abs/2011.01535, 2020.

\bibitem{DBLP:conf/iccvw/GansbekeBNPG19}
Wouter~Van Gansbeke, Bert~De Brabandere, Davy Neven, Marc Proesmans, and
  Luc~Van Gool.
\newblock End-to-end lane detection through differentiable least-squares
  fitting.
\newblock In {\em 2019 {IEEE/CVF} International Conference on Computer Vision
  Workshops, {ICCV} Workshops 2019, Seoul, Korea (South), October 27-28, 2019},
  pages 905--913. {IEEE}, 2019.

\bibitem{DBLP:conf/iccv/GarnettCPLL19}
Noa Garnett, Rafi Cohen, Tomer Pe'er, Roee Lahav, and Dan Levi.
\newblock 3d-lanenet: End-to-end 3d multiple lane detection.
\newblock In {\em 2019 {IEEE/CVF} International Conference on Computer Vision,
  {ICCV} 2019, Seoul, Korea (South), October 27 - November 2, 2019}, pages
  2921--2930. {IEEE}, 2019.

\bibitem{DBLP:journals/corr/abs-2109-12218}
Jake Grigsby, Zhe Wang, and Yanjun Qi.
\newblock Long-range transformers for dynamic spatiotemporal forecasting.
\newblock {\em CoRR}, abs/2109.12218, 2021.

\bibitem{hendy2020fishing}
Noureldin Hendy, Cooper Sloan, Feng Tian, Pengfei Duan, Nick Charchut, Yuesong
  Xie, Chuang Wang, and James Philbin.
\newblock Fishing net: Future inference of semantic heatmaps in grids.
\newblock {\em arXiv preprint arXiv:2006.09917}, 2020.

\bibitem{DBLP:conf/cvpr/HomayounfarMLU18}
Namdar Homayounfar, Wei{-}Chiu Ma, Shrinidhi~Kowshika Lakshmikanth, and Raquel
  Urtasun.
\newblock Hierarchical recurrent attention networks for structured online maps.
\newblock In {\em 2018 {IEEE} Conference on Computer Vision and Pattern
  Recognition, {CVPR} 2018, Salt Lake City, UT, USA, June 18-22, 2018}, pages
  3417--3426. {IEEE} Computer Society, 2018.

\bibitem{homayounfar2018hierarchical}
Namdar Homayounfar, Wei-Chiu Ma, Shrinidhi~Kowshika Lakshmikanth, and Raquel
  Urtasun.
\newblock Hierarchical recurrent attention networks for structured online maps.
\newblock In {\em Proceedings of the IEEE Conference on Computer Vision and
  Pattern Recognition}, pages 3417--3426, 2018.

\bibitem{homayounfar2019dagmapper}
Namdar Homayounfar, Wei-Chiu Ma, Justin Liang, Xinyu Wu, Jack Fan, and Raquel
  Urtasun.
\newblock Dagmapper: Learning to map by discovering lane topology.
\newblock In {\em Proceedings of the IEEE/CVF International Conference on
  Computer Vision}, pages 2911--2920, 2019.

\bibitem{hong2019rules}
Joey Hong, Benjamin Sapp, and James Philbin.
\newblock Rules of the road: Predicting driving behavior with a convolutional
  model of semantic interactions.
\newblock In {\em Proceedings of the IEEE Conference on Computer Vision and
  Pattern Recognition}, pages 8454--8462, 2019.

\bibitem{jaritz20202d}
Maximilian Jaritz.
\newblock {\em 2D-3D scene understanding for autonomous driving}.
\newblock PhD thesis, PSL Research University, 2020.

\bibitem{DBLP:journals/corr/abs-2002-06604}
YeongMin Ko, Jiwon Jun, Donghwuy Ko, and Moongu Jeon.
\newblock Key points estimation and point instance segmentation approach for
  lane detection.
\newblock {\em CoRR}, abs/2002.06604, 2020.

\bibitem{DBLP:conf/icra/LiWWZ22}
Qi Li, Yue Wang, Yilun Wang, and Hang Zhao.
\newblock Hdmapnet: An online {HD} map construction and evaluation framework.
\newblock In {\em 2022 International Conference on Robotics and Automation,
  {ICRA} 2022, Philadelphia, PA, USA, May 23-27, 2022}, pages 4628--4634.
  {IEEE}, 2022.

\bibitem{liang2019convolutional}
Justin Liang, Namdar Homayounfar, Wei-Chiu Ma, Shenlong Wang, and Raquel
  Urtasun.
\newblock Convolutional recurrent network for road boundary extraction.
\newblock In {\em Proceedings of the IEEE/CVF Conference on Computer Vision and
  Pattern Recognition}, pages 9512--9521, 2019.

\bibitem{liang2018end}
Justin Liang and Raquel Urtasun.
\newblock End-to-end deep structured models for drawing crosswalks.
\newblock In {\em Proceedings of the European Conference on Computer Vision
  (ECCV)}, pages 396--412, 2018.

\bibitem{ma2019exploiting}
Wei-Chiu Ma, Ignacio Tartavull, Ioan~Andrei B{\^a}rsan, Shenlong Wang, Min Bai,
  Gellert Mattyus, Namdar Homayounfar, Shrinidhi~Kowshika Lakshmikanth, Andrei
  Pokrovsky, and Raquel Urtasun.
\newblock Exploiting sparse semantic hd maps for self-driving vehicle
  localization.
\newblock {\em arXiv preprint arXiv:1908.03274}, 2019.

\bibitem{DBLP:conf/ivs/NevenBGPG18}
Davy Neven, Bert~De Brabandere, Stamatios Georgoulis, Marc Proesmans, and
  Luc~Van Gool.
\newblock Towards end-to-end lane detection: an instance segmentation approach.
\newblock In {\em 2018 {IEEE} Intelligent Vehicles Symposium, {IV} 2018,
  Changshu, Suzhou, China, June 26-30, 2018}, pages 286--291. {IEEE}, 2018.

\bibitem{pan2020cross}
Bowen Pan, Jiankai Sun, Ho~Yin~Tiga Leung, Alex Andonian, and Bolei Zhou.
\newblock Cross-view semantic segmentation for sensing surroundings.
\newblock {\em IEEE Robotics and Automation Letters}, 5(3):4867--4873, 2020.

\bibitem{paz2021tridentnet}
David Paz, Hengyuan Zhang, and Henrik~I Christensen.
\newblock Tridentnet: A conditional generative model for dynamic trajectory
  generation.
\newblock {\em arXiv preprint arXiv:2101.06374}, 2021.

\bibitem{paz2020probabilistic}
David Paz, Hengyuan Zhang, Qinru Li, Hao Xiang, and Henrik Christensen.
\newblock Probabilistic semantic mapping for urban autonomous driving
  applications.
\newblock {\em arXiv preprint arXiv:2006.04894}, 2020.

\bibitem{philion2020lift}
Jonah Philion and Sanja Fidler.
\newblock Lift, splat, shoot: Encoding images from arbitrary camera rigs by
  implicitly unprojecting to 3d.
\newblock In {\em Proceedings of the European Conference on Computer Vision},
  2020.

\bibitem{ravi2018real}
B Ravi~Kiran, Luis Roldao, Benat Irastorza, Renzo Verastegui, Sebastian Suss,
  Senthil Yogamani, Victor Talpaert, Alexandre Lepoutre, and Guillaume Trehard.
\newblock Real-time dynamic object detection for autonomous driving using prior
  3d-maps.
\newblock In {\em Proceedings of the European Conference on Computer Vision
  (ECCV) Workshops}, pages 0--0, 2018.

\bibitem{rella2021decoder}
Edoardo~Mello Rella, Jan-Nico Zaech, Alexander Liniger, and Luc Van~Gool.
\newblock Decoder fusion rnn: Context and interaction aware decoders for
  trajectory prediction.
\newblock {\em arXiv preprint arXiv:2108.05814}, 2021.

\bibitem{richards1999remote}
John~Alan Richards and JA Richards.
\newblock {\em Remote sensing digital image analysis}, volume~3.
\newblock Springer, 1999.

\bibitem{DBLP:conf/cvpr/RoddickC20}
Thomas Roddick and Roberto Cipolla.
\newblock Predicting semantic map representations from images using pyramid
  occupancy networks.
\newblock In {\em 2020 {IEEE/CVF} Conference on Computer Vision and Pattern
  Recognition, {CVPR} 2020, Seattle, WA, USA, June 13-19, 2020}, pages
  11135--11144. {IEEE}, 2020.

\bibitem{seif2016autonomous}
Heiko~G Seif and Xiaolong Hu.
\newblock Autonomous driving in the icity—hd maps as a key challenge of the
  automotive industry.
\newblock {\em Engineering}, 2(2):159--162, 2016.

\bibitem{sun2019leveraging}
Tao Sun, Zonglin Di, Pengyu Che, Chun Liu, and Yin Wang.
\newblock Leveraging crowdsourced gps data for road extraction from aerial
  imagery.
\newblock In {\em Proceedings of the IEEE/CVF Conference on Computer Vision and
  Pattern Recognition}, pages 7509--7518, 2019.

\bibitem{ventura2018iterative}
Carles Ventura, Jordi Pont-Tuset, Sergi Caelles, Kevis-Kokitsi Maninis, and Luc
  Van~Gool.
\newblock Iterative deep learning for road topology extraction.
\newblock {\em arXiv preprint arXiv:1808.09814}, 2018.

\bibitem{wu2020motionnet}
Pengxiang Wu, Siheng Chen, and Dimitris~N Metaxas.
\newblock Motionnet: Joint perception and motion prediction for autonomous
  driving based on bird's eye view maps.
\newblock In {\em Proceedings of the IEEE/CVF Conference on Computer Vision and
  Pattern Recognition}, pages 11385--11395, 2020.

\bibitem{yang2018hdnet}
Bin Yang, Ming Liang, and Raquel Urtasun.
\newblock Hdnet: Exploiting hd maps for 3d object detection.
\newblock In {\em Conference on Robot Learning}, pages 146--155. PMLR, 2018.

\bibitem{DBLP:conf/siu/YenIaydinS18}
Yasin YenIaydin and Klaus~Werner Schmidt.
\newblock A lane detection algorithm based on reliable lane markings.
\newblock In {\em 26th Signal Processing and Communications Applications
  Conference, {SIU} 2018, Izmir, Turkey, May 2-5, 2018}, pages 1--4. {IEEE},
  2018.

\bibitem{DBLP:conf/eccv/YuMRZY20}
Cunjun Yu, Xiao Ma, Jiawei Ren, Haiyu Zhao, and Shuai Yi.
\newblock Spatio-temporal graph transformer networks for pedestrian trajectory
  prediction.
\newblock In Andrea Vedaldi, Horst Bischof, Thomas Brox, and Jan{-}Michael
  Frahm, editors, {\em Computer Vision - {ECCV} 2020 - 16th European
  Conference, Glasgow, UK, August 23-28, 2020, Proceedings, Part {XII}}, volume
  12357 of {\em Lecture Notes in Computer Science}, pages 507--523. Springer,
  2020.

\bibitem{zaech2020action}
Jan-Nico Zaech, Dengxin Dai, Alexander Liniger, and Luc Van~Gool.
\newblock Action sequence predictions of vehicles in urban environments using
  map and social context.
\newblock In {\em International Conference on Intelligent Robots and Systems
  (IROS)}, 2020.

\bibitem{zhou2022cross}
Brady Zhou and Philipp Kr{\"a}henb{\"u}hl.
\newblock Cross-view transformers for real-time map-view semantic segmentation.
\newblock In {\em Proceedings of the IEEE/CVF Conference on Computer Vision and
  Pattern Recognition}, pages 13760--13769, 2022.

\end{thebibliography}
}

\end{document}